# Correlated Action Effects in Decision Theoretic Regression

**Craig Boutilier**
Department of Computer Science
University of British Columbia
Vancouver, BC, CANADA, V6T 1Z4
**email:** cebly@cs.ubc.ca

## Abstract

Much recent research in decision theoretic planning has adopted Markov decision processes (MDPs) as the model of choice, and has attempted to make their solution more tractable by exploiting problem structure. One particular algorithm, *structured policy construction* achieves this by means of a decision theoretic analog of *goal regression*, using action descriptions based on Bayesian networks with tree-structured conditional probability tables. The algorithm as presented is not able to deal with actions with correlated effects. We describe a new decision theoretic regression operator that corrects this weakness. While conceptually straightforward, this extension requires a somewhat more complicated technical approach.

## 1 Introduction

While Markov decision processes (MDPs) have proven to be useful as conceptual and computational models for decision theoretic planning (DTP), there has been considerable effort devoted within the AI community to enhancing the computational power of these models. One of the key drawbacks of classic algorithms such as policy iteration [13] or value iteration [2] is the need to explicitly "sweep through" state space some number of times to determine the values of various actions at different states. Because state spaces grow exponentially with the number of features relevant to the problem description, such methods are wildly impractical for realistic planning problems, a difficulty dubbed by Bellman the "curse of dimensionality."

Recent research on the use of MDPs for DTP has focussed on methods for solving MDPs that avoid explicit enumeration of the state space while constructing optimal or approximately optimal policies. Such techniques include the use of *reachability analysis* to eliminate (approximately) unreachable states [9, 1], and *state aggregation*, whereby various states are grouped together and each aggregate state or "cluster" is treated as a single state. Recently, methods for automatic aggregation have been developed in which certain problem features are ignored, making certain states indistinguishable [8, 3, 11, 5, 16].

In some of these aggregation techniques, the use of standard AI representations like STRIPS or Bayesian networks to represent actions in an MDP can be exploited to help construct the aggregations. In particular, they can be used to help identify which variables are *relevant*, at any point in the computation of an optimal policy, to the determination of value or to the choice of action. This connection has lead to the insight that the basic operations in computing optimal policies for MDPs can be viewed as a generalization of *goal regression* [5]. More specifically, a *Bellman backup* [2] for a specific action $a$ is essentially a regression step where, instead of determining the the conditions under which one specific goal proposition will be achieved when $a$ is executed, we determine the conditions under which $a$ will lead to a number of different "goal regions" (each having different value) such that the probability of reaching any such goal region is fixed by the conditions so determined. Any set of conditions so determined for action $a$ is such that the states having those conditions all accord the same expected value to the performance of $a$. The net result of this *decision theoretic regression operator* is a partitioning of state space into regions that assign different expected value to $a$. Classical goal regression can be viewed as a special case of this, where the action is deterministic and the value distinction is binary (goal states versus nongoal states).

A decision theoretic regression operator of this form is developed in [5]. The value functions being regressed are represented using decision trees, and the actions that are regressed through are represented using Bayes nets with tree-structured conditional probability tables. As shown there (see also [4]), classic algorithms for solving MDPs, such as value iteration or modified policy iteration, can be expressed purely in terms of decision theoretic regression, together with some tree manipulation. Unfortunately, the particular algorithm presented there assumes that actions effects are uncorrelated, imposing a restriction on the types of Bayes nets that can be used to represent actions.[1] The aim of this paper is to correct this deficiency. Specifically, we describe the details of a decision theoretic regression al-gorithm that handles such correlations in the effects of actions and the difficulties that must be dealt with. We note that this paper does not offer much in the way of a conceptual advance in the understanding of the decision theoretic regression, and builds directly on the observations in [5, 4]. However, the modifications of these approaches to handle correlations are substantial enough, both in technical detail and in spirit, to warrant special attention.

[1] Correlated effects can be represented by clustering variables, of course, but such a representation is often unnatural and can cause a substantial blowup in representation size.

We review MDPs and their representation using Bayes nets and decision trees in Section 2. We briefly describe the basic decision theoretic regression operator of [5] in Section 3. In Section 4, we illustrate the challenges posed by correlated action effects for decision theoretic regression with several examples and describe an algorithm that meets these challenges. We conclude in Section 5 with some remarks on future research and related work.

## 2 MDPs and Their Representation

### 2.1 Markov Decision Processes

We assume that the system to be controlled can be described as a fully-observable, discrete state *Markov decision process* [2, 13, 15], with a finite set of system states $S$. The controlling agent has available a finite set of actions $A$ which cause stochastic state transitions: we write $\Pr(s, a, t)$ to denote the probability action $a$ causes a transition to state $t$ when executed in state $s$. A real-valued reward function $R$ reflects the objectives of the agent, with $R(s)$ denoting the (immediate) utility of being in state $s$.[2] A (stationary) *policy* $\pi : S \to A$ denotes a particular course of action to be adopted by an agent, with $\pi(s)$ being the action to be executed whenever the agent finds itself in state $s$. We assume an infinite horizon (i.e., the agent will act indefinitely) and that the agent accumulates the rewards associated with the states it enters.

In order to compare policies, we adopt *expected total discounted reward* as our optimality criterion; future rewards are discounted by rate $0 \leq \beta < 1$. The value of a policy $\pi$ can be shown to satisfy [13]:

$$V_\pi(s) = R(s) + \beta \sum_{t \in S} Pr(s, \pi(s), t) \cdot V_\pi(t)$$

The value of $\pi$ at any initial state $s$ can be computed by solving this system of linear equations. A policy $\pi$ is *optimal* if $V_\pi(s) \geq V_{\pi'}(s)$ for all $s \in S$ and policies $\pi'$. The *optimal value function* $V^*$ is the same as the value function for any optimal policy.

A number of techniques for constructing optimal policies exist. An especially simple algorithm is *value iteration* [2]. We produce a sequence of *n-step optimal value functions* $V^n$ by setting $V^0 = R$, and defining

$$V^{i+1}(s) = \max_{a \in \mathcal{A}} \{R(s) + \beta \sum_{t \in S} Pr(s, a, t) \cdot V^i(t)\} \quad (1)$$

[2] More general formulations of reward (e.g., adding action costs) offer no special complications.

The sequence of functions $V^i$ converges linearly to $V^*$ in the limit. Each iteration is known as a *Bellman backup*. After some finite number $n$ of iterations, the choice of maximizing action for each $s$ forms an optimal policy $\pi$ and $V^n$ approximates its value.

There are several variations one can perform on the Bellman backup. For instance, given a policy $\pi$, we can compute the value of $\pi$ by means of *successive approximation*. If we set $V_\pi^0 = R$, and define

$$V_\pi^{i+1}(s) = \{R(s) + \beta \sum_{t \in S} Pr(s, \pi(s), t) \cdot V_\pi^i(t)\} \quad (2)$$

Then $V_\pi^k(s)$ denotes the expected value of performing $\pi$, starting at $s$, for $k$ steps; this quantity converges to $V_\pi(s)$. Finally, given a value function $V$, we define the *Q-function* [17], mapping state-action pairs into values, as follows:

$$Q(s, a) = \{R(s) + \beta \sum_{t \in S} Pr(s, a, t) \cdot V(t)\} \quad (3)$$

This denotes the value of performing action $a$ at state $s$ assuming that value $V$ is attained at future states (e.g., if we acted optimally after performing $a$ and attained $V^*$ subsequently). We use $Q_a$ to denote the $Q$-function for a particular action $a$ (i.e., $Q_a(s) = Q(s, a)$). It is not hard to see that value iteration and successive approximation can be implemented by repeated construction of $Q$-functions (using the current value function), and the appropriate selection of $Q$-values (either by maximization at a particular state, or by using the policy to dictate the correct action and $Q$-value to apply to a state).

### 2.2 Action and Reward Representation

One of the key problems facing researchers regarding the use of MDPs for DTP is the "curse of dimensionality:" the number of states grows exponentially with the number of problem variables. Since the representation of transition probabilities, reward and value functions, policies, as well as the computations involved in dynamic programming algorithms, all involve enumerating states, the representation of MDPs and the computational requirements of solution techniques can be quite onerous. Fortunately, several good representations for MDPs, suitable for DTP, have been proposed. These include stochastic STRIPS operators [14, 3] and dynamic Bayes nets [10, 5]. We will use the latter.

We assume that a set of variables $\mathbf{V}$ describes our system. To represent actions and their transition probabilities, for each action we have a *dynamic Bayes net* (DBN) with one set of nodes representing the system state prior to the action (one node for each variable), another set representing the world after the action has been performed, and directed arcs representing causal influences between these sets. Our convention is to use the notation $X'$ to denote that variable $X$ after the occurrence of the action and $X$ to denote $X$ before the action. Each post-action node has an associated *conditional probability table* (CPT) quantifying the influence of the action on the corresponding variable, given the value of its influences (see [5, 7] for a more detailed discussion of this representation).[3] Figures 1(a) and (b) illustrate this representation for two different actions. We use $\Pi(X')$ to denote the parents of node $X'$ in a network and $val(X)$ to denote the values variables $X$ (or $X'$) can take.

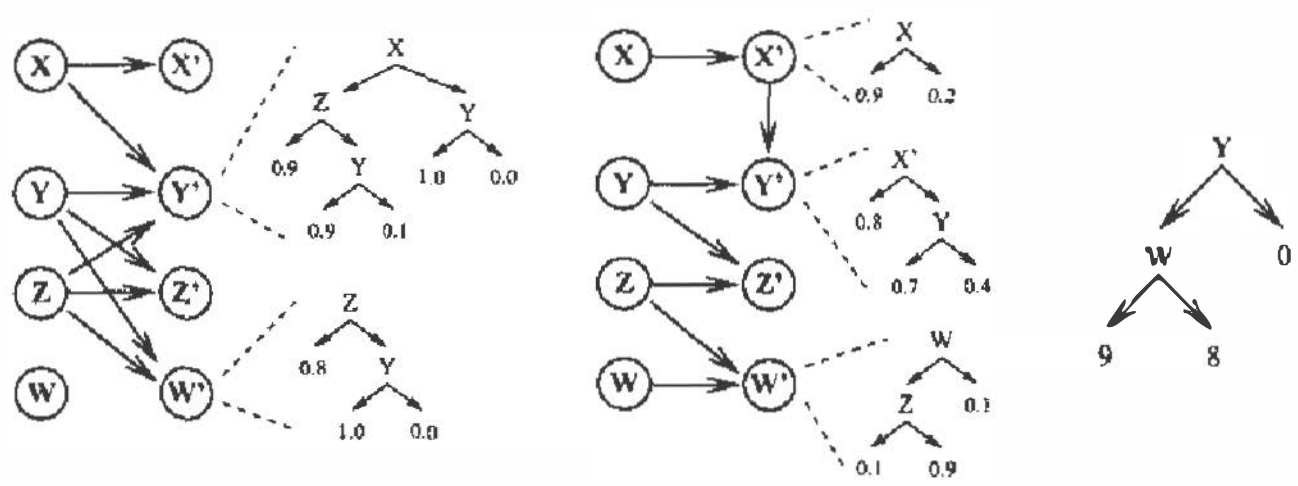

Figure 1: (a) Action network (no correlations); (b) Action network (correlations); and (c) Reward Tree

The lack of an arc from a pre-action variable $X$ to a post-action variable $Y'$ in the network for action $a$ reflects the independence of $a$'s effect on $Y$ from the prior value of $X$. We capture additional independence by assuming structured CPTs; that is, we exploit *context-specific independence* (CSI) as defined in [6]. In particular, we use a *decision tree* to represent the function that maps parent variable values to (conditional) probabilities. For instance, the trees in Figure 1(a) show that $Z$ influences the probability of $Y$ becoming true (as a consequence of the action), but only if $X$ is true (left arrows are assumed to be labeled "true" and right arrows "false"). We refer to the tree-structured CPT for node $X'$ in the network for action $a$ as $Tree(X', a)$. We make special note of the existence of the arc between $X'$ and $Y'$ in Figure 1(b). This indicates that the effect of action $a$ on $X$ and $Y$ is *correlated*. We will see that such arcs pose challenges for decision theoretic regression.

Finally, a decision tree can also be used to represent the reward function $R$, as shown in Figure 1(c). We call this the (immediate) reward tree, $Tree(R)$. We will also use this representation for value functions and Q-functions.

## 3 Decision Theoretic Regression: Uncorrelated Effects

Apart from the naturalness and conciseness of representation offered by DBNs and decision trees, these representations lay bare a number of regularities and independencies that can be exploited in optimal and approximate policy construction. Methods for optimal policy construction can use compact representations of policies and value functions in order to prevent enumeration of the state space.

In [5], a structured version of modified policy iteration is developed, in which value functions and policies are represented using decision trees and the DBN representation of the MDP is exploited to build these compact policies.[4] This technique is applied in [4] to value iteration, and dynamic approximation methods are considered as well. Roughly, if one has a tree representation of a value function, only certain variables will be mentioned as being relevant (under certain conditions) to value. When performing Bellman backups, the fact that certain variables are irrelevant to, say, $V^n$ means that action-condition pairs that are distinguished by their influence on irrelevant variables need not be distinguished in the representation or computation of $V^{n+1}$.

The key to all of these algorithms is a *decision theoretic regression operator* used to construct the $Q$-functions for an action $a$ given a specific value function. If the value function is tree-structured, this algorithm produces a *$Q$-tree*, a tree-structured representation of the $Q$-function that obviates the need to compute $Q$-values on a state-by-state basis. We note that since: (a) the initial value function $Tree(R)$ is tree-structured; (b) the algorithm for producing $Q$-trees retains this structure; and (c) the algorithm for "merging" $Q$-trees (e.g., by maximization) also retains this structure; then the resulting value function will be structured (and methods for building structured policies based on this can be easily defined). We focus here only on the construction of $Q$-trees—the remaining parts of the algorithms are straightforward. As in [5, 4], we assume that no action has correlated effects (all have the form illustrated in Figure 1(a)): this simplifies the algorithm considerably.

Let $a$ be the action described in Figure 1(a), and let the tree in Figure 1(c) correspond to some value function $V$ (call it $Tree(V)$). To produce the Q-function $Q_a$ based on $V$ according to Equation 3, we need to determine that the probabilities with which different states $s$ make the conditions dictated by the branches of $Tree(V)$ true.[5] It should be clear, since $a$'s effects on the variables in $Tree(V)$ exhibit certain regularities (as dictated by its network), that $Q_a$ should also exhibit certain regularities. These are discovered in the following algorithm for constructing a Q-tree representing (the future value component of) $Q_a$ given $Tree(V)$ and a network for $a$.

1. Generate an ordering $O_V$ of variables in $Tree(V)$.
2. Set $Tree(Q_a) = \emptyset$
3. For each variable $X$ in $Tree(V)$ (using ordering $O_V$):
   (a) Determine contexts $c$ in $Tree(V)$ (partial branches) that lead to an occurrence of $X$.
   (b) At any leaf of $Tree(Q_a)$ such that $\Pr(c) > 0$ for some context $c$: replace the leaf with a copy of $Tree(X, a)$ at that leaf (retain $\Pr(X)$ at each leaf of $Tree(X, a)$); remove any redundant nodes from this copy; for each $Y$ ordered before $X$ such that $\Pr(Y)$ labeled this leaf of $Tree(Q_a)$, copy $\Pr(Y)$ to each leaf of $Tree(X, a)$ just added.

[3] To simplify the presentation, we restric our attention to binary variables in our examples.

[4] See [12] for a similar, though less general, method in the context of reinforcement learning, where deterministic action effects and specific goal regions are assumed).

[5] We ignore the fact that states with different reward have different Q-values; these differences can be added easily once the future reward component of Equation 3 has been spelled out.

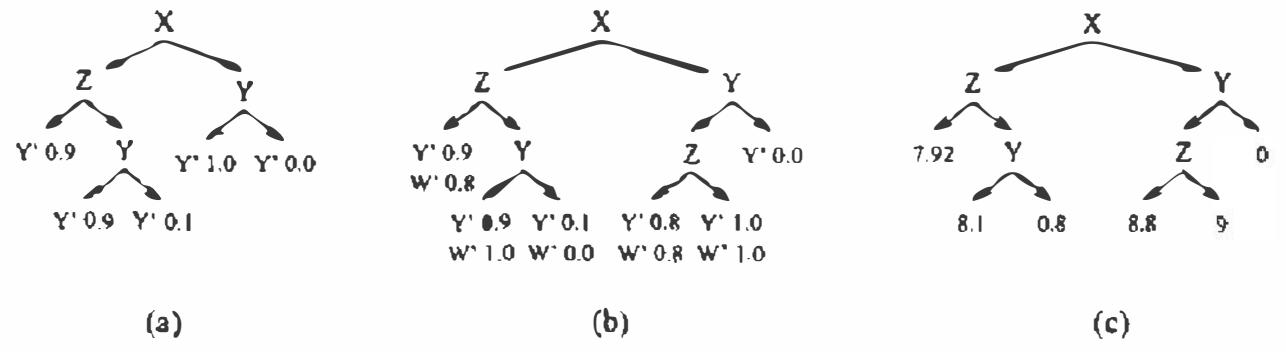

Figure 2: Decision theoretic regression: no correlations

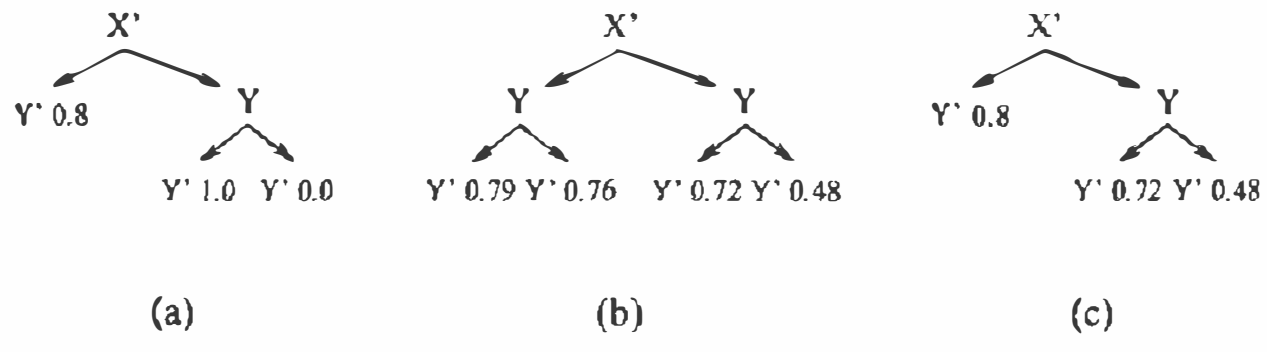

Figure 3: Dec. theoretic regression: summing over parents

4. At each leaf of *Tree*($Q_a$), replace probabilities labeling leaf with $\sum_c \Pr(c)V(c)$, using these probabilities to determine $\Pr(c)$ for any context (branch) of *Tree*($V$).

This decision theoretic regression algorithm forms the core of the policy construction techniques of [5, 4].

We illustrate the algorithm on the example above. We will *regress* the variables of *Tree*($V$) through action $a$ in the order $Y, W$ (generally, we want to respect the ordering within the tree as much as possible). We first regress $Y$ through $a$, producing the tree shown in Figure 2(a). Notice that this tree accurately reflects $\Pr(Y')$ when $a$ is executed given that the previous state satisfies the conditions labeling the branches. We then regress $W$ through $a$ and add the results to any branch of the tree so far where $\Pr(Y) > 0$ (see Figure 2(b)). Thus, *Tree*($W, a$) is not added to the rightmost branch of the tree in Figure 2(a)—this is because if $Y$ is known to be false, $W$ has no impact on reward, as dictated by *Tree*($V$). Notice also that because of certain redundancies in the tests (internal nodes) of *Tree*($Y, a$) and *Tree*($W, a$), certain portions of *Tree*($W, a$) can be deleted. Figure 2(b) now accurately describes the probabilities of both $Y$ and $W$ given that $a$ is executed under the listed conditions, and thus dictates the probability of making any branch of *Tree*($V$) true: we simply multiply $\Pr(W)$ and $\Pr(Y)$ for the values of $W$ and $Y$ labeling this branch. Therefore, the (future component of the) expected value of performing $a$ at any such state can easily be computed at each leaf of this tree using $\sum\{\Pr(c)V(c) : c \in branches(Tree(V))\}$—the result is shown in Figure 2(c).

It is important to note that the justification for this very simple algorithm lies in the fact that, in the network for $a$, $Y'$ and $W'$ are independent given any context $k$ labeling a branch of *Tree*($Q_a$). This ensures that the term $\Pr(Y'|k)\Pr(W'|k)$ corresponds to $\Pr(Y', W'|k)$. There are two reasons for this. First, since no action effects are correlated, the effect of $a$ on any variable is independent given knowledge of the previous state (i.e., the post-action variables are independent given the pre-action variables). Second, this independence does not require complete knowledge of the state, but can exploit both the variable independence specified by the network structure, and the CSI relations dictated by the CPTs.

## 4 Regression with Correlated Action Effects

As noted above, the fact that action effects are uncorrelated means that knowledge of the previous state renders all post-action variables independent. This is not the case when effects are correlated as in Figure 1(b). This can lead to several difficulties for decision theoretic regression. The first is the fact that, although we want to compute the expected value of $a$ given only the state $s$ of pre-action variables, the probability of post-action variables that can influence value (e.g., $Y'$) is not specified solely in terms of the pre-action state, but also involves other post-action variables (e.g., $X'$). This difficulty is relatively straightforward to deal with, requiring that we sum out the influence of post-action variables on other post-action variables.

The second problem requires more sophistication. Because action effects are correlated, the probability of the variables in *Tree*($V$) may also be correlated. This means that determining the probability of attaining a certain branch of *Tree*($V$) by considering the "independent" probabilities of attaining the variables on the branch (as in the previous section) is doomed to failure. For instance, if both $X$ and $Y$ lie on a single branch of *Tree*($V$), we cannot compute $\Pr(X'|s)$ and $\Pr(Y'|s)$ independently to determine the probability $\Pr(X', Y'|s)$ of attaining that branch. To deal with this, we must construct Q-trees where the *joint distribution* over certain subsets of variables is computed.

We illustrate the necessary intuitions behind a new algorithm for decision theoretic regression that adequately deals with correlations (i.e., arbitrary DBNs) through a series of examples. We then present the algorithm in its entirety.

### 4.1 Summing out Post-Action Influences

Consider action $a$ in Figure 1(b) and *Tree*($V$) in Figure 1(c). Using the algorithm from the previous section to produce *Tree*($Q_a$), we would first regress $Y'$ through $a$ to obtain the tree shown in Figure 3(a). Continuation of the algorithm will not lead to a legitimate Q-tree, since it involves a post-action variable $X'$. Our revised algorithm will establish the dependence of $\Pr(Y')$ on previous state $s$ by "summing out" the influence of $X'$ on $Y'$, letting $Y'$ vary with the parents of $X'$. Specifically, we will simply compute

$$\begin{aligned}\Pr(Y'|s) &= \sum_{x' \in val(X')} \Pr(Y'|x', s) \cdot \Pr(x'|X) \\ &= \sum_{x' \in val(X')} \Pr(Y'|x', Y) \cdot \Pr(x'|X)\end{aligned}$$

This will proceed as follows. Once we have regressed $Y'$ through $a$, we will replace the node $X'$ by *Tree*($X', a$). This dictates $\Pr(X'|\Pi(X'))$. Denote the subtree of the replaced node corresponding to each values $x_i$ of $X'$ by $STree(x_i)$. Now at each leaf $l$ of $Tree(X', a)$ just added, we have recorded $\Pr(x_i)$. For those values of $x_i$ that have positive probability, we merge the trees $STree(x_i)$ and copy these at $l$.[6] In Figure 3(b), we have placed the merged subtree rooted at $Y$ under both $X = x$ and $X = \overline{x}$. Now, at each leaf we can determine (indeed, we have recorded while building the tree) both $\Pr(X'|X)$ and $\Pr(Y'|X', Y)$ for the appropriate values of $X$ and $Y$ labeling the branch. We can then compute $\Pr(Y)$ as needed, depending only on pre-action variables. Once completed, it is easy to see that regression of $W'$ through $a$ can proceed unhindered as in the last section.

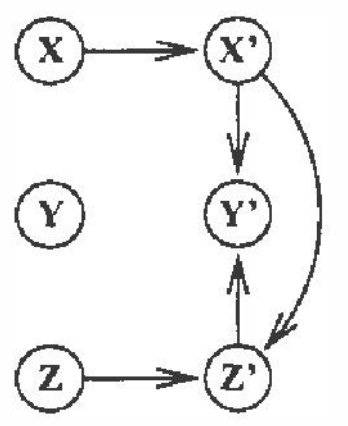

Figure 4: Ordering variables for replacement

We note that had the CPT for $X'$ indicated that $\Pr(x'|x) = 1$ (instead of 0.9), we would not have copied the $X' = \overline{x}'$ subtree under $X = x$. This is because the influence of $Y$ on $Y'$ is only valid when $X'$ is false. The result would have been the simpler tree shown in Figure 3(c). Finally, we see that had there been a chain of dependence among post-action variables, this replacement of post-action variables in the regressed tree by their parents can simply proceed recursively. For instance, had $X'$ depended on a third variable $V'$, this variable would have been introduced with $Tree(X', a)$. The influence of $V'$ on $Y'$ could then have been summed out in a similar fashion.

We now consider a second example (see Figure 4) that illustrates that the order in which these post-action variables are replaced in a tree can be crucial. Suppose that we have an action $a$ similar to the one just described, except now we have that variable $Y'$ depends on both $X'$ and $Z'$ (i.e., $a$'s effect on $X$, $Y$ and $Z$ is correlated). When we regress $Y'$ through $a$, we will introduce a tree in which both $X'$ and $Z'$ appear, and we assume that $X'$ and $Z'$ appear together on at least one branch of $Tree(Y', a)$ that is present in $Tree(Q_a)$. Now let us suppose that $Z'$ also depends on $X'$, as in Figure 4. In such a case, it is important to substitute $Tree(Z', a)$ for $Z'$ before substituting $Tree(X', a)$ for $X'$. If we replace $X'$ first, we will compute

$$\Pr(Y'|Z', \Pi(X')) = \sum_{x' \in val(X')} \Pr(Y'|x', Z') \Pr(x'|\Pi(X'))$$

(we suppress mention of other parents of $Y'$). Subsequently, we would replace occurrences of $Z'$ with $Tree(Z')$ and compute

$$\Pr(Y'|\Pi(Z'), \Pi(X')) = \sum_{z' \in val(Z')} \Pr(Y'|z', \Pi(X')) \Pr(z'|\Pi(Z'))$$

This ordering has two problems. First, since $X'$ is a parent of $Z'$, this approach would reintroduce $X'$ into the tree, requiring the wasted computation of summing out $X'$ again. Even worse, for any branch of $Tree(Z')$ on which $X'$ occurs, the computation above is not valid, for $Y'$ is not independent of $X'$ (an element of $\Pi(Z')$) given $Z'$ and $\Pi(X')$ (since $X'$ directly influences $Y'$).

Because of this, we require that when a variable $Y'$ is regressed through $a$, if any two of its post-action parents lie on the same branch of $Tree(Y')$, these nodes in $Tree(Y')$ must be replaced by their trees in an order that respects the dependence among post-action variables in $a$'s network. More precisely, let a *post-action ordering* $O_P$ for action $a$ be any ordering of variables such that, if $X'$ is a parent of $Z'$, then $Z'$ occurs *before* $X'$ in this ordering (so the ordering goes against the direction of the within-slice arcs). Post-action variables in $Tree(Y')$, or any tree obtained by recursive replacement of post-action variables, must be replaced according to some post-action ordering $O_P$.

## 4.2 Computing Local Joint Distributions

Consider again $Tree(V)$ shown in Figure 1(c) and its regression through the action $a$ shown in Figure 5(a). Figure 5(b) shows the regression of $Y'$ through $a$. We would normally then insert $Tree(W', a)$ at each leaf of this tree, and replace the $Y'$ node of this tree with $Tree(Y', a)$. Of course, $\Pr(Y')$ already labels each leaf, so we can immediately replace the node $Y'$ in $Tree(W', a)$ with its merged subtrees (as described in the previous subsection).[7] The structure of this tree is indicated in Figure 5(c). If we were to proceed as above, we would simply sum out the influence of $Y'$ on $W'$ to determine $\Pr(W')$ at each leaf. That is, we compute

$$\Pr(W'|W, X, Y) = \sum_{y' \in val(Y')} \Pr(W'|y', W) \cdot \Pr(y'|X, Y)$$

This, unfortunately, does not provide an accurate picture of the probability of attaining the conditions $c$ labeling the branches of $Tree(V)$. If we labeled the leaves of the tree in Figure 5(c) with $\Pr(Y')$ and $\Pr(W')$ so computed, these probabilities, while correct, are not sufficient to determine $\Pr(Y', W')$: $Y'$ and $W'$ are not independent given $X, Y, W$. Instead, we need the joint distribution $\Pr(Y', W')$ labeling the leaves, as shown in Figure 5(c). We note that this joint is obtained in a very simple fashion. At each leaf we have recorded $\Pr(Y')$ and $\Pr(W'|Y')$ (under the appropriate conditions). Instead of summing out

[6] Merging simply requires creating a tree whose branches make the distinction contained in each subtree. We do this by ordering the trees, and grafting each tree in order onto the leaves of the tree resulting from merging it predecessor, and removing redundant nodes (i.e., duplicated tests) as appropriate.

[7] And as described above, we do not need to include the $Y' = \overline{y}$ subtree (from $Tree(W', a)$) at the $X = x$ leaf, since $\Pr(Y') = 1$ labels that leaf.

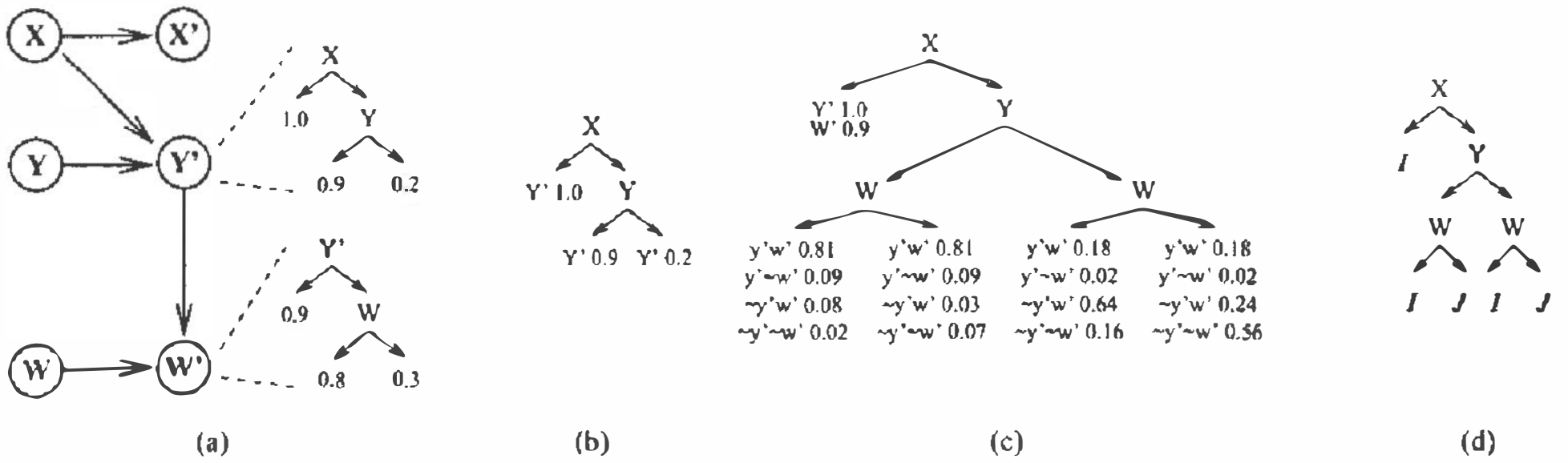

Figure 5: Decision theoretic regression: correlated reward variables

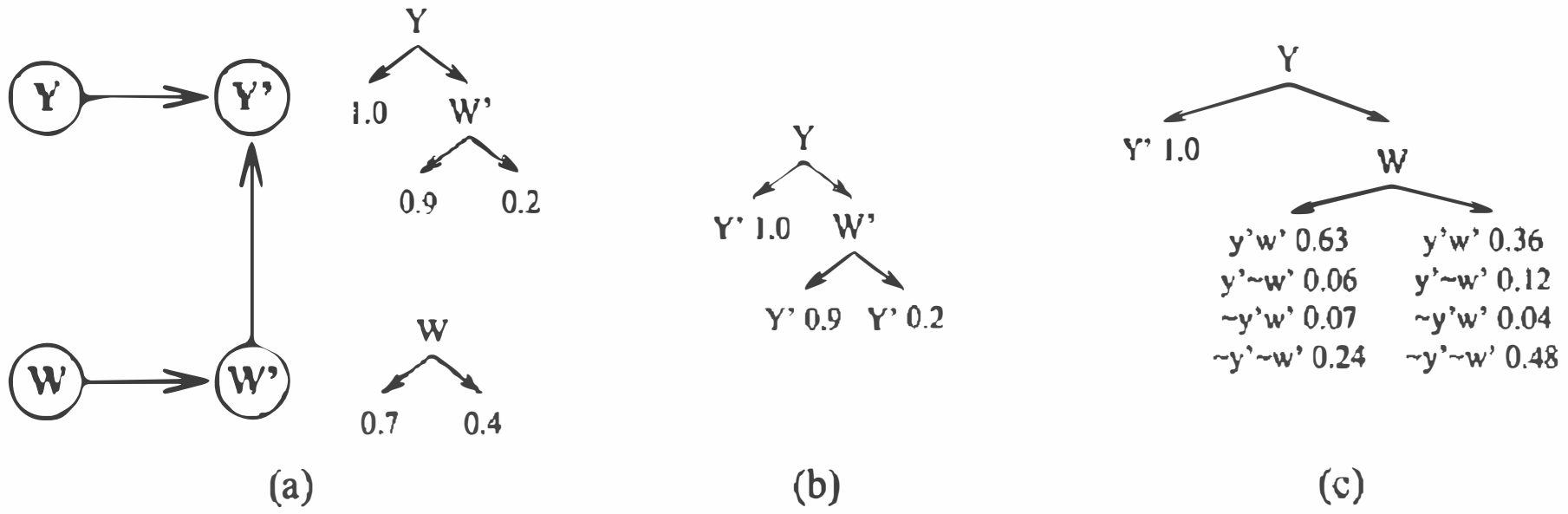

Figure 6: Decision theoretic regression: correlated reward variables

the influence of $Y'$ on $W'$, we explicitly store the terms $\Pr(Y', W')$ we compute.[8]

This approach, explicitly representing the joint probability of different action effects instead of summing out the influence of in-slice parents, allows us to accurately capture the correlations among action effects that directly impact the value function. We need only compute the joint distribution between two relevant variables in contexts in which they are actually correlated. For instance, suppose that we switched the locations of variables $Y'$ and $W$ in *Tree*$(W', a)$ in Figure 5(a). We see then that $W'$ only depends on $Y'$ when $W$ is false. In this case, the final regressed tree (before expected value is computed) would have a similar shape, as shown in Figure 5(d); but we would compute joints only at the $\overline{w}$-leaves (labeled $J$). Independent probabilities for $Y'$ and $W'$ can be computed and stored in the usual fashion at the other leaves (labeled $I$).

The last piece in the puzzle pertains to the decision of when to sum out a variable's influence on an in-slice descendent and when to retain the (local) joint representation. Consider the usual value tree and the action $a$ shown in Figure 6(a); notice that the dependence of $W'$ on $Y'$ has been reversed.

Regression of $Y'$ leads to the tree in Figure 6(b). When removing the influence of variable $W'$ on $Y'$, we obtain the tree shown in Figure 6(c). Using the usual ideas from above, we would be tempted to sum out the influence of $W'$ on $Y'$, computing

$$\Pr(Y'|Y, W) = \sum_{w' \in val(W')} \Pr(Y'|w', Y) \cdot \Pr(w'|W)$$

However, if we "look ahead," we see that we will later have to regress $W'$ at both leaf nodes for which we are attempting to compute $\Pr(Y')$. Clearly, since these are correlated, we should leave $\Pr(Y')$ uncomputed (explicitly), leaving the joint representation of $\Pr(Y', W')$ as shown in Figure 6(c). When subsequently regressing $W'$ at each leaf where $\Pr(Y') > 0$, our work is already done at these points.

This leads to an obvious question: when removing a post-action variable $V'$ from the tree produced when regressing another variable $Y'$ which depends on it, under what circumstances should we sum out the influence of $V'$ on $Y'$ or retain the explicit joint representation of $\Pr(V', Y')$? Intuitively, we want to retain the "expansion" of $Y'$ in terms of $V'$ (i.e., retain the joint) if we will *need* to worry about the correlation between $Y'$ and $V'$ later on. As we saw above, this notion of *need* is easily noticed when one of the variables in directly involved in the value tree, and will be regressed explicitly afterward (under the conditions that label the current branch of course). However, variables that may be *needed* subsequently are not restricted to those that have to be regressed directly (i.e., they needn't be part of *Tree*$(V)$); instead, variables that *influence* those in *Tree*$(V)$ can sometimes be retained in expanded form.

[8] We should emphasize that this local joint distribution does not need to be computed or represented explicitly. Any *factored* representation, e.g., storing directly $\Pr(Y')$ and $\Pr(W'|Y')$, can be used. In fact, when a number of variables are correlated, we generally expect this to be the approach of choice. However, we will continue to speak as if the local joint were explicitly represented for ease of exposition.

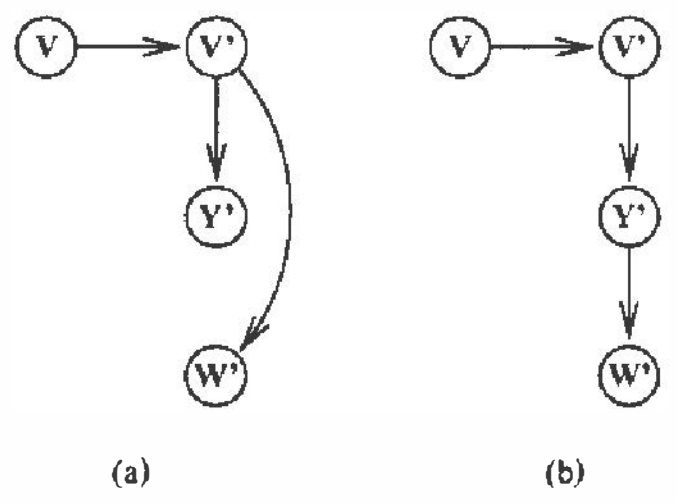

Figure 7: Detecting future need for parents

Consider the action in Figure 7(a) (we again use the usual *Tree*$(V)$). When we regress $Y'$ through $a$, we obtain a tree containing node $V'$, which subsequently gets replaced by *Tree*$(V', a)$. The term $\Pr(Y')$ should be computed explicitly by summing the terms $\Pr(Y'|v') \cdot \Pr(v'|V)$ over values $v'$. However, looking at *Tree*$(V)$, we see that $W'$ will be regressed wherever $\Pr(Y') > 0$, and that $W'$ also depends on $V'$. This means that (ignoring any CSI) $W'$ and $Y'$ are correlated given the previous state $s$. This dependence is mediated by $V'$, so we will need to explicitly use the joint probability $\Pr(Y', V')$ to determine the joint probability $\Pr(Y', W')$. In such a case, we say that $V'$ is *needed* and we do not sum out its influence on $Y'$. In an example like this, however, once we have determined $\Pr(Y', V', W')$ we can decide to sum out $V'$ if it won't be needed further.

Finally, suppose that $W'$ depends indirectly on $V'$, but that this dependence is mediated by $Y'$, as in Figure 7(b). In this case, we can sum out $V'$ and claim that $V'$ is *not needed*: $V'$ can only influence $W'$ through its effect on $Y'$. This effect is adequately summarized by $\Pr(Y'|V)$; and the terms $\Pr(Y', V'|V)$ are not needed to compute $\Pr(Y', W'|V)$ since $W'$ and $V'$ are independent given $Y'$. We provide a formal definition of *need* in the next section.

### 4.3 An Algorithm for Decision Theoretic Regression

The intuitions illustrated by the previous examples can be put together in an algorithm. We assume that an action $a$ in network form has been provided with tree-structured CPTs (that is, *Tree*$(X', a)$ for each post-action variable $X'$), as well as a value tree *Tree*$(V)$. We let $O_V$ be an ordering of the variables within *Tree*$(V)$, and $O_P$ some post-action ordering for $a$. The following algorithm constructs a Q-tree for $Q_a$ with respect to *Tree*$(V)$.

1. Set *Tree*$(Q_a) = \emptyset$
2. For each variable $X$ in *Tree*$(V)$ (using $O_V$):
   (a) Determine contexts $c$ in *Tree*$(V)$ (partial branches) that lead to an occurrence of $X$.
   (b) At any leaf $l$ of *Tree*$(Q_a)$ such that $\Pr(c) > 0$ for some context $c$, add *simplify*(*Tree*$(X', a), l, k)$ to $l$, where $k$ is the context in *Tree*$(Q_a)$ leading to $l$ (we treat $l$ as its label).
3. At each leaf of *Tree*$(Q_a)$, replace the probability terms (of which some may be joint probabilities) labeling the leaf with $\sum_c \Pr(c)V(c)$, using these probabilities to determine $\Pr(c)$ for any context (branch) of *Tree*$(V)$.

The key intuitions described in our earlier examples are part of the algorithm that produces *simplify*(*Tree*$(X', a), l, k)$. Recall that $l$ is a leaf of the current (partial) *Tree*$(Q_a)$ and is labeled with (possibly joint) probabilities of some subset of the variables in *Tree*$(V)$. Context $k$ is the set of conditions under which those probabilities are valid; note that $k$ can only consist of pre-action variables. Simplification involves the repeated replacement of the post-action variables in *Tree*$(V)$ and the recording of joint distributions if required. It proceeds as follows:

1. *Reduce* *Tree*$(X', a)$ for context $k$ by deleting redundant nodes.
2. For any variables $Y'$ in *Tree*$(X', a)$ whose probability is part of the label for $l$, *replace* $Y'$ in *Tree*$(X', a)$, respecting the ordering $O_P$ in replacement. That is, for each occurrence of $Y'$ in *Tree*$(X', a)$:
   (a) merge the subtrees under $Y'$ corresponding to values $y$ of $Y'$ that have positive probability, deleting $Y'$;
   (b) compute $\Pr(X'|Y', m) \cdot \Pr(Y')$ for each leaf in the merged subtree (let this leaf correspond to context $m = k \wedge k'$, where $k'$ is the branch through *Tree*$(X', a)$);
   (c) if $Y'$ has been regressed at $l$ or is *needed* in context $m$, label this leaf of the merged tree with the joint distribution over $X', Y'$; otherwise, sum out the influence of $Y'$.
3. For any remaining variables $Y'$ in *Tree*$(X', a)$, *replace* $Y'$ in *Tree*$(X', a)$, respecting ordering $O_P$ in replacement; i.e.,
   (a) replace each occurrence of $Y'$ with *Tree*$(Y', a)$ (and reduce by context $n = k \wedge k'$, where $k'$ is the branch through *Tree*$(X', a)$ leading to $Y'$);
   (b) to each leaf $l'$ of the *Tree*$(Y', a)$ just added, merge the subtrees under $Y'$ corresponding to values $y$ of $Y'$ that have positive probability at $l'$;
   (c) proceed as in Step (2).
4. Repeat Step 3 until all new post-action variables introduced at each iteration of Step 3 have been removed. For any variable removed from the tree, we construct a joint distribution with $X'$ if it is *needed*, or sum over its value if it is not.

These steps embody the intuitions described earlier. We note that when we refer to $\Pr(Y')$ as it exists in the tree, it may be that $\Pr(Y')$ does not label the leaf explicitly but jointly with one or more other variables. In such a case, when we say that $\Pr(X', Y')$ should be computed, or $Y'$ should be summed out, we intend that $X'$ will become part of the explicit joint involving other variables. Any variables that are part of such a cluster are correlated with $Y'$ and hence with $X'$. Variables can be summed out once they are no longer needed.

The last requirement is a formal definition of the concept of *need*—as described above, this determines when to retain a joint representation for a post-action variable that is being removed from *Tree*$(Q_a)$. Let $l$ be the label of the leaf where $X'$ is being regressed, $k$ be the context leading to that leaf, $Y'$ be the ancestor of $X'$ being replaced, and $k'$ the context labeling the branch through (partially replaced) *Tree*$(X', a)$ where the decision to compute $\Pr(X')$ or $\Pr(X', Y')$ is being made. We say that $Y'$ is *needed* if:

1. there is a branch $b$ of *Tree*$(V)$ on which $Y'$ lies, such that $b$ has positive probability given $l$; or

2. there is a branch $b$ on which $Z$ lies, such that $b$ has positive probability given $l$; $\Pr(Z')$ is not recorded in $l$; and there is a path from $Y'$ to $Z'$ in $a$'s network that is *not* blocked by $\{X', k, k'\}$.

## 5 Concluding Remarks

We have presented an algorithm for the construction of Q-trees using a Bayes net representation of an action, with tree-structured CPTs, and a tree-structured value function. This forms the core of a decision theoretic regression algorithm. Unlike earlier approaches to this problem, this algorithm works with arbitrary Bayes net action descriptions, and is not hindered by the presence of "intra-slice" arcs in the network reflecting correlated action effects. This is an important feature because this representational power allows one to concisely represent actions in a natural fashion. Forcing someone to specify actions without correlations is often unnatural, and the translation into a network with no intra-slice arcs (e.g., by clustering variables) can cause a blowup in the network size and the inability to exploit many independencies in decision theoretic regression.

One concern about such approaches is the overhead involved in constructing appropriate trees. We note that this algorithm will behave exactly as the algorithms discussed in [5, 4] if there are no correlations. While we expect MDPs to often contain actions that exhibit correlations, it seems likely that many of these correlations will be localized. Furthermore, the use of context-specific independence allows clustering to be performed only under the *specific* conditions that give rise to the dependencies among effects. Finally, we observe that we are only concerned with *maintaining* correlations among variables that actually influence value. If we are dealing with effects that impact other relevant effects, but are not of direct interest themselves, these are summed out immediately with little overhead.

We are currently exploring the extent to which networks can be preprocessed to alleviate some of the repeated operations at different regression steps. There is also an interesting connection to the recent work of Michael Littman (personal communication); he has suggested the transformation of action representations such as ours into a STRIPS representation of actions that does not require correlated effects to be represented explicitly. This is achieved by a radical transformation of the problem, but one that is polytime, requires only a polynomial size increase, and from which an optimal policy can be extracted in polynomial time. It is an open question if that method exploits the same type of structural regularities as our approach. Finally, we hope to consider the use of other compact CPT representations in decision theoretic regression.

### Acknowledgements

This research was supported by NSERC Research Grant OGP0121843 and IRIS NCE Program IC-7.